# Faster Convolution Inference Through Using Pre-Calculated Lookup Tables

Grigor Gatchev, M.D.
Technical University
Sofia, Bulgaria
grigor@gatchev.info

Valentin S. Mollov, Ph.D.
Technical University
Sofia, Bulgaria
mollov_vs@yahoo.com

*Abstract: Low-cardinality activations permit an algorithm based on fetching the inference values from pre-calculated lookup tables instead of calculating them every time. This algorithm can have extensions, some of which offer abilities beyond those of the currently used algorithms. It also allows for a simpler and more effective CNN-specialized hardware.*

## INTRODUCTION

The convolutional neural networks (CNNs) are among the most used types of artificial neural networks (ANNs). However, they require an amount of computing resources that makes them unfeasible for many applications. Lowering this requirement can extend their usability.

V. Sze et al.[15] categorize the existing research on the efficient processing in ANNs into three classes, based on their design levels: hardware platforms, memory technologies and software algorithms. We overview some of the research in these categories and discuss it.

Based on that, we propose a convolutional inference algorithm, suitable for low-cardinality activations. Its essence is fetching the inference values from pre-calculated lookup tables instead of calculating them every time. It can achieve higher productivity and can work on simpler and faster specialized hardware than most CNN algorithms.

## RECENT RESEARCH REPORTS

### Hardware Platforms

J. Schmidthuber states that the advances in the ANNs during the last 20 years are made possible by the advances in hardware. Some of the more recent ones are:

- Google created Tensor Processing Unit[20], an ASIC accelerator implemented as a matrix multiplication engine, controlled by a host processor. Currently has three generations; a fourth is being developed. An embedded-targeting version of it is Edge TPU. Jouppi et al.[35] find in 2016 that TPU performance exceeds by 15-30-80x that of contemporary CPUs and GPUs.

- NVidia develops the Tensor Core technology, based on their 3D video accelerator technology, and used in their DGX[21] workstations and servers.

- Intel offer the Movidius Myriad Vision Processing Unit[34], featuring DRAM, imaging / vision accelerators, an array of VLIW vector processors called SHAVE processors and a SPARC CPU core.

- CEVA[44] offers CEVA-XM6, a vision and deep learning processing unit featuring VPU architecture, and NeuPro-S, adding on top of CEVA-XM specialized engines for convolution, activation and pooling.

- Synopsys[45] has DesignWare EV Vision Processor series, combining scalar core with DSP and a programmable CNN engine.

- Cadence[46] offers the Tensilica customizable-by-client CNN vision processors, able to combine many modules widely used in CNN ICs.

- Qadeer et al.[43] present Convolution Engine, one of the first CNN-specialized ASICs, which is 8-15x faster than a SIMD processor.

- Ardakani et al.[23] propose a computational method inspired by a computational core of fully connected networks, and implement it in CMOS, showing up to 9.5 times faster work than the accelerators reported up to 2017.

- Zhao et al.[24] created a block parallel computing algorithm based on MTCA. It allows high parallel hardware implementation, and saves up to 82% storage space, compared to the img2col approach.

- Chen et al.[41] design an accelerator for large-scale CNNs and deep neural networks (DNNs) that is 117x faster and uses 21x less energy than a 128-bit 2GHz SIMD processor. Chen et al.[25] extend this architecture to a multi-chip system with on-chip storage as a specialized supercomputer. Du et al.[42] place such an accelerator next to a sensor, improving efficiency by eliminating DRAM access.

- Wang et al.[26] design an energy-efficient architecture for binary weight neural networks (BWNN) and demonstrate that it achieves efficiency over 2.0TOp/s/W when scaled to 65 nm – twice that of art prior to 2018.

- Pullini et al.[36] propose a hybrid HW/SW CNN accelerator, targeted to wearable and IoT scenarios, implemented as 65-nm system-on-chip, integrating a near-threshold parallel processor cluster and a hardware accelerator for convolution-accumulation operations.

- Cavigelli et al.[37] present an optimized convolutional network implementation, suitable for real-time scene labeling on embedded platforms, running on the Nvidia Tegra K1 embedded SoC. Part of the same team[40] created Origami, a CNN accelerator with reduced bandwidth requirements and better efficiency than most previous works.

- Chen et al.[39] implement with Eyeriss v2 an entire network-on-chip, using flexible hierarchical mesh, able to run both compact and sparse DNNs.

- Moons et al.[47] show an energy-efficient dynamic precision scalable processor for ConvNets with 256 parallel units, able to exploit convolutions sparsity.

## Memory Technologies

The memory technology with most promise at efficiency appears to be the usage of integer activations with low bit width instead of floating-point ones. It decreases both the memory usage (and therefore the RAM cache misses) and the calculating power required. Many papers report no or negligible precision loss with these. Some examples are:

- Daily[2] finds that INT8 operations, compared to FP32 ones, achieve energy consumption savings of 30x for addition and 18.5x for multiplication, and on-chip area savings of 116x for addition and 27x for multiplication.

- Ilin et al.[5] apply 8-bit fixed-point arithmetic (INT8) for approximate calculations in image recognition.

- Integer-Net by Truong et al.[14] reports 7x reduction of the memory footprint in comparison with a 32-bit floating-point (FP32) implementation, while having only 2% worse recognition performance.

- Wu et al.[16] use in their WAGE model integer weights, activations, gradients and error, sized 2, 8, 8 and 8 bits respectively.

- Das et al.[3] use 16-bit (INT16) and 32-bit (INT32) integers as dynamic fixed point values in implementations of AlexNet and other NNs, achieving the accuracy of the originals with improved throughput on Xeon CPUs.

- de Bruin et al.[1] implement NNs on constrained hardware (embedded ARM CPUs) by achieving sufficient quantization on 16-bit CPU accumulators.

- F. Zhu et al.[19] research gradients in unified 8-bit integer training of NNs and propose universal techniques for managing it that avoid the direction deviation and illegal gradient updates.

- Jacob et al.[6] provide a scheme that quantizes weights and activations as INT8, and bias vectors as INT32.

- Lin et al.[11] use an integer adder instead of floating point multiplier in IA-Net, targeting memory reduction through model compression, and also achieve 20% reduction of the inference time.

- Rastegari et al.[13] propose Binary Weight Networks, where filter weights have width of 1 bit, and XNOR-Networks, where both weights and layer activations have width of 1 bit, and achieve results equal to or better than AlexNet and a BinaryNet implementation of ImageNet.

- Li et al.[10], in an attempt to improve the precision of the Binary Weight Networks, propose Ternary Weight Networks, where weights can have three possible values instead of only two. Zhu et al.[18] add quantization to these.

- Lin et al[12] try to improve the accuracy of CNNs with binary weights and activations by approximating full-precision weights with linear combination of multiple binary weight bases, and using multiple binary activations to alleviate information loss.

- Gysel et al.[4] develop a framework for approximating neural networks with reduced bit width versions of its values, which is often able to fine-tune a network to using INT8 values with a loss of precision smaller than 1%.

## Software Algorithms

The original convolutional direct multiplication algorithm (DM) uses a big number of multiplications that comprises most of the processing load in a CNN. Due to this, it is a key target for algorithmic improvements. Some recently suggested ones are:

- Mathieu at al.[27] compute convolutions as Fourier pointwise products while reusing the same transformed feature map, achieving speedup of over a magnitude.

- Highlander et al.[28] increase convolution speed by using fast Fourier transform (FFT).

- Nguyen-Thanh et al.[29] use 2D-FFT-based algorithm to lower energy consumption, creating conditions for processing speedup.

- Abtahi et al.[30] improve convolution speed several times by using FFT variants.

- Lin et al.[31] use tile-based decomposition with Fourier transforms to create a fast convolution algorithm called tFFT.

- Chitsaz et al.[32] find that splitting is an effective solution of some problems in FFT computation with small kernels, eg. in a typical CNN.

- Lavin et al.[22] suggest a family of fast matrix multiplication algorithms, based on Winograd's minimal filtering, which can reduce the CNN multiplications by a factor of 2.25.

- Liu et al.[33] note that ANNs using Winograd- and FFT-based convolution algorithms cannot handle well NN compression, but can be pruned to 10% and 25% of their original size respectively, with only 0.18% loss of accuracy.

- Ju et al.[49] analyze most popular fast convolution algorithms as formal bilinear ones, and show that overlap-add and Winograd family algorithms rival the accuracy of FFT without using complex arithmetic. They give a corollary for the minimum bilinear algorithm rank of a linear convolution, and present algorithms that achieve it. They also present algorithms that minimize the convolution error.

- Kim et al.[51] find that in AlexNet on GPUs both FFT and Winograd / Toom-Cook methods achieve speedup of up to 4x over the DM method. However, this difference also is expected to decrease on custom ASICs, due to the bigger and more complex circuitry required.

- Jin et al.[79] use one-dimensional filters (flattened / factorized networks) to emulate 3-D convolutional filters without performance loss. Wang et al.[80] use similar type of factorization, combined with topological connections.

- Sifre[78] introduces separable convolution. Lebedev et al.[74] use spatially separable convolution to increase productivity with only a minor reduction of precision.

- Chollet[75] and Ghosh[76] avoid some limitations of the spatially separable convolution by using depthwise convolution.

- Fialka et al.[50] note that in the typical CNN scenario (a small filter being convolved with large data) applying filters by separable convolutions is much faster than FFT, possibly due to the constant factors of the latter (use of complex arithmetic).

- Lebedev et al.[74] and He et al.[77] achieve further speedup with separable convolutions by applying decomposition (CP- and depth-wise, respectively).

- Habib et al.[81] explore the different techniques for accelerating the CNN training, and focus on the usage of stochastic gradient descent.

- Han et al.[48] decrease the memory requirements of NNs by 35-49x without accuracy loss by introducing three-stage "deep compression": connection pruning, weights quantization and subsequent Huffman coding.

- Thoma[82] develops a method for visualization of classification errors with confusion matrices, and bases on it hierarchical classifiers.

- Young et al.[83] compress CNN weights post-training through transform quantization. Wu et al.[84] do the same through quantization based on scale optimization.

## DISCUSSION

### Hardware Tendencies

The development of widely used computation-intensive information processing solutions appears to follow a common tendency. They usually start as CPU-only software implementations, then move the computationally intensive operations to GPUs, then to FPGAs, and then to ASICs. Finally, they start shifting to the ASICs as much of the other solution functionalities as their complexity allows.

The potential for applied usage of CNNs, alone or as a component of complex ANNs, appears to be huge – from supercomputers emulating powerful general intelligence, to narrowly specialized embedded solutions. They appear to undergo this evolution too. Most passed the CPU-only stage, currently use as accelerators GPUs and specialized matrix computation hardware, and some experimental ones run on FPGAs. It appears only reasonable to assume that they will eventually move to ASICs. These will likely start as CNN-specific accelerators that enhance CPUs, and will gradually acquire most, if possible – all of the CNN data processing.

There are already attempts to create multi-purpose or combined NN SoC processors and ASICs (eg. D. Shin et al. DNPU[38]). However, the computing requirements of CNNs and most other widely used NN types (fewer weights and more multiply/add operations) are too different. This inevitably leads to creating ICs that are effective for only one of the supported NN types, or contain different on-chip modules for every supported type (leading to big on-chip size), or have a lot of configuration circuitry and lower effectiveness than a type-specific NN ASIC would.

In addition, production tasks that would require dynamically switching an NN processor between types of NN processing are extremely rare; usually a production task will use it for only one type of processing. Using for a task an ASIC that is customized to the appropriate NN type (eg. CNN), or even to the specific CNN subtype, can bring better productivity. It can already be safely predicted that specialized ASICs will soon find many applications in mass-produced items, which will justify their development.

Consequently, the research in the area of new memory technologies and software algorithms would likely benefit from taking this evolution into account – namely the fact that they will eventually be used on custom ASICs. Some of the things to consider are:

- On modern CPUs and GPUs, the execution time for corresponding INT and FP operations is the same or similar. However, their energy consumption and the on-chip area and complexity of their underlying circuitry differs by more than a magnitude[2]. So, on specialized ASICs their speed will likely differ by a similar ratio.

- In specialized ASICs, limiting data size decreases the on-chip size and the energy consumption, and increases the speed. So, limiting the INT size will be beneficial where a satisfactory precision can be maintained.

- Using booleans instead of INT values can simplify the on-chip circuitry to a significant degree, if a satisfactory precision can be achieved. For example, if one of two values to be multiplied is a boolean, a specialized ASIC can utilize an adder instead of a multiplier, lowering the time, the energy consumption and the on-chip size.

- Simpler algorithms need simpler circuitry in a specialized ASIC, thus resulting potentially in faster work and lower energy consumption. For example, the DM algorithm can prove in some cases (eg. certain tasks performed on highly optimized ASICs) more productive than some theoretically faster but much more complex algorithms, eg. FFT or Winograd / Toom-Cook ones.

- Some operations naturally require much simpler and smaller on-chip circuitry, and / or are executed much faster on a specialized ASIC. For example, bit shifting and masking perform much better than multiplication and division, or even addition and subtraction.

### Memory Considerations

As noted in the Hardware Tendencies section, smaller-size values not only require less memory. They also enable achieving better performance, smaller energy consumption and smaller on-chip size. Due to this, limiting the values size is an important way to higher productivity.

Activations and weights of smaller cardinality tend to produce results with higher granularity. This might make the network easier to analyze, and to allow achieving better compression or pruning. This in turn might additionally improve the performance.

The key problem here is that the increased granularity of the weights and the activations can decrease the precision below the acceptable level. One of the ways to check that appears to be to investigate how the biological neural networks (BNNs) work in this aspect.

#### *Activation cardinality in BNNs*

Research is conducted and methods are proposed for estimating the maximal and the typical amount of information that a biological neuron or neuronal population is able to convey ([52], [53], [54], [55], [56], [57], [58], [59]). It often shows an amount of information that would require in an equivalent ANN significant activations cardinality, sometimes up to INT32 or FP32, or even beyond that.

However, this is not always the same as the amount of information passed between two neurons. The differing signal characteristics in an axon and in a chemical synapse limit the maximum of information that can be transmitted consecutively through both per unit of time:

- Most biological neurons that process information are spiking neurons. In these, different characteristics of neuronal spikes (edge, form, height, length, frequency, distance from surrounding spikes etc) are usually merged in a chemical synapse into one characteristic – amount of mediator released in the synaptic cleft per time. In most cases, the only characteristics contributing significantly to that are the spikes height, length and frequency.

- Some of the characteristics, like time difference, overlap, count of synapses, their proximity to the neuron body, level of the synaptic mediator in the neuron environment etc, affect in most (but not all) neurons their activation only through some kind of summation. In ANNs that is all reflected through the neuron input weights and their control.

- In most spiking neuron types, spikes in a type of neurons almost always have the same height and duration. As the mediator released into the synaptic cleft is quickly removed, its present amount is effectively determined by the frequency of the spikes ("rate coding"[60]).

- Due to the quick removal, the amount of mediator released in the synaptic cleft per time must exceed certain level in order to to reach the post-synaptic membrane and affect the next neuron. This puts a lower boundary to the spikes frequency that can affect the next neuron through a synapse. Also, the refractory period after a spike creates an upper boundary to the spikes frequency that a neuron can achieve. These form the neuron's usual spikes frequency range.

- In extreme conditions many neurons can exceed this range, but few have a useful role outside of it. A neuron's effective work mode is usually within their usual range, and in most neurons covers only a part of it.

- Different types of neurons can have very different durations of activation. Those that process quickly changing information (QCIP neurons), like that in most ANNs and virtually all CNNs, tend to have very short activation duration. Even at a maximal frequency of spikes, only a limited number of spikes can fit in an activation. In a synapse, this number is determined by the neuron with the shorter activation time.

All this limits the cardinality needed to express the typical and the maximal amount of information passed between two neurons. In most types of QCIP neurons, the maximal amount of information can be expressed with INT4, and for some types is a boolean value.

Neurons can vary with the time the number of synapses connecting two of them. However, this does not change the information cardinality that a single activation can cover. It is expressed in ANNs through adjusting the input weights.

Some biological neuronal types use temporal coding, phase of firing, population coding and other means to convey information, in addition to or instead of rate coding ([61], [62], [63], [64], [65], [66], [67], [68], [69], [70], [71]). However, these can usually be best, sometimes only matched in ANNs not by larger activation cardinality, but by using specific kinds of artificial neurons, eg. LSTMs.

The other type of information-transmitting junction between neurons – the gap junction[91] – is not analyzed here, due to uncertainties (as of 2021) about some details of its role in information processing[87], [90]. It appears to be important in some places (retina etc[86], [88]), but to play relatively small role in the information processing in most BNNs (except for modulating chemical synapses[85]). The currently accepted opinion is that where signal transmission is concerned, it effectively connects multiple neurons into one. Further research might change this opinion, and might bring new ideas into ANNs.

If most QCIP neurons and their structures in the BNNs can provide the results they are able to deliver through usage of activations with effective 4-bit cardinality or lower, it can be concluded that many CNNs too should be able to achieve that. This conclusion appears to be supported by some experimental results ([10], [13], [14], [18]).

In addition, big ANNs that do complex information processing would often be optimally implemented not as entirely programmable structures. Significant parts of them would likely be partially or completely hardwired structures that can only change their weights and possibly their connectivity to a degree. Some of these structures might need big activation cardinality and / or usage of specific algorithms. Many however will work well with small activation cardinality and standardized algorithms. This includes stacks of convolutional layers and layers that typically support them – effectively CNN networks.

This section concludes that in tasks similar to those performed by many BNNs, many CNNs would be able to perform adequately with activation cardinality up to INT4, and relatively few would need activation cardinality over INT8.

*Weights cardinality in BNNs*

The cardinality of the weight analogue in a BNN depends on the number of synapses between a neuron and its afferent (input), their position (ones closer to the soma usually have stronger effect) and their strength. The number and strength of the synapses are shown to be modified by the learning process[93], [94], and both common logic and preliminary data show that strengthening of a connection may result in creating synapses close to the some of the efferent neuron.

Some types of neurons can have up to several thousands of synapses with an afferent[92]. The strength of a synapse and its closeness to the soma are analog parameters that cannot be quantized reliably. So, the maximal possible weights cardinality in BNNs appears to need at least INT20, and maybe beyond INT32.

However, very few biological neurons appear to actually have analogues to such big weights – mostly ones involved in very fine output regulation, depending on big amount of inputs with high signal cardinality (eg. the Purkinje cells in the mammal cerebellum). Neurons integrating few inputs, including some in multistage range-limiting information processing areas, appear to work well with much smaller range for their weight analogues. For example, neurons that have about up to 20 afferent synapses appear to usually have weights analogue with cardinality up to INT8. Many neurons in the mammal brain fall within this category.

This section concludes that in tasks similar to those performed by BNNs, some CNN elements might need input weights cardinality up to INT32, or even FP32, but many would not need input weights cardinality beyond INT8. Thus, CNN structures that support different-sized input weights might be able to work in down to almost ¼ of the memory that would be required by CNN structures with uniformly sized weights. So, algorithms that allow for the former can bring a significant memory economy.

## Algorithm Discussions

Most of the proposed more productive algorithms and algorithm optimizations are based on mathematical methods for optimizing matrix multiplications. While this is the most obvious venue for improving the productivity, it appears to not be the only one. So, investigating methods outside of it appears promising.

FFTs are shown by many authors to bring a productivity increase. However, they also have their problems, and most proposed solutions to those increase further their algorithm complexity in CNNs.

The Winograd / Toom-Cook family of algorithms avoids some of the FFT problems, while being significantly simpler and producing networks similarly suitable to post-learning compression. However, they still suffer from similar loss of precision, compared to the DM algorithm.

The separable convolution algorithms offer substantial reduction of the multiplications needed – in some cases over 10x. However, in both spatially separable and depth-wise convolution this also comes with substantial reduction of the number of network parameters. This might limit the result precision and thus be inappropriate for some tasks.

The O-notation formula of most of these improvements – FFT, Karatsuba / Winograd / Toom-Cook algorithms, linear convolution, separable convolution – is commendable. However, in highly optimized CNN ASICs their complexity might slow them and increase the on-chip size and the energy consumption of the circuitry needed by them. This can make them impractical.

A direction that attracts efforts during the last few years is the CNN optimization during and after learning. Different variants of network decomposition, pruning and other methods for compression are described, showing good productivity improvements and minimal, often negligible loss of precision. Consequently, CNN algorithms that produce more optimizable networks show greater promise.

### Discussions Summary

Most attempts to increase the productivity of the CNNs tend to fall strictly within one of the three major classes by Sze et al. – hardware, memory and algorithms. They rarely cross the borders of a single class, and even when they do, it most often appears to be a collateral benefit rather than a pre-planned cross-class design.

At the same time, it appears that seeking optimizations that tailor solutions to more than one class might produce better results than concentrating on one class only.

Improvements research outside of these three classes also appears nearly nonexistent. Due to being almost unexplored, investigating this area might produce interesting results. For example, newest BNN knowledge appears to be underused in latest NN research, despite providing valuable insights.

### THE PCILT ALGORITHM

Here is proposed an algorithm, targeted at custom CNN ASICs. It can be executed on simpler and faster hardware than most widely accepted algorithms.

It is appropriate in CNNs that use activations with small cardinality, and are constrained primarily by computing power, or prioritize productivity over memory economy. There it is faster than the classic DM algorithm.

### Basic Version

Prior to the learning start, the multiplications of the filter values by all possible activation values are calculated and placed in pre-calculated inference lookup tables (PCILTs) (Fig. 1). While making an inference, the convolutional function results are obtained from those tables, instead of calculating them each time (Fig. 2).

The main advantage of the algorithm is better productivity. In the classical convolution algorithm, two values (data / activation value and filter value) are obtained from memory and then multiplied. This algorithm instead treats the first obtained value (the data / activation value) as an offset in a PCILT table, and obtains the multiplication result from memory as the table value at this offset. This eliminates the multiplication operation.

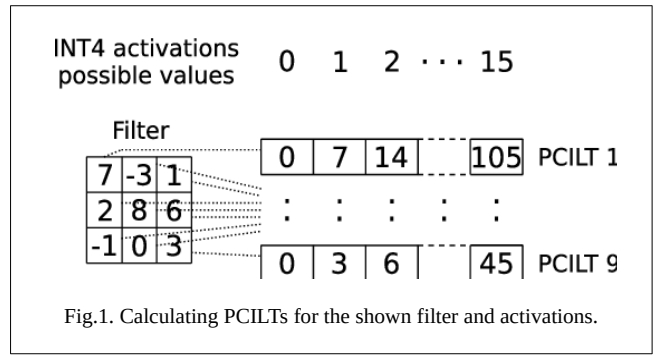

Fig.1. Calculating PCILTs for the shown filter and activations.

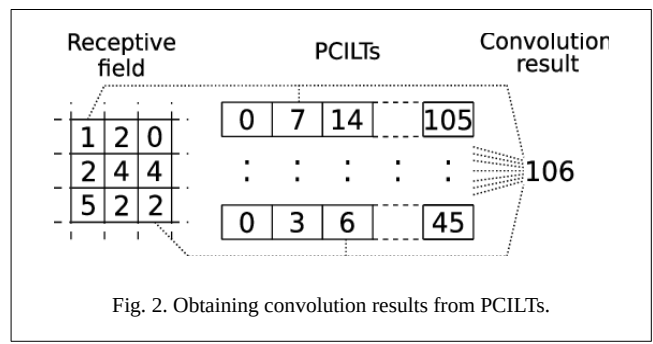

Fig. 2. Obtaining convolution results from PCILTs.

In a modern CPU obtaining a memory value is a slower operation than multiplication of data in registers, esp. if that memory is not already in the CPU memory cache. There, the speed advantage of the basic PCILT algorithm is expected to be small, if any. In a custom ASIC however, a PCILT can be implemented as a fast memory block, having its own address and data buses, situated next to the results adder. It would use the activation value as an address and pass the value at this address to the adder (Fig. 3).

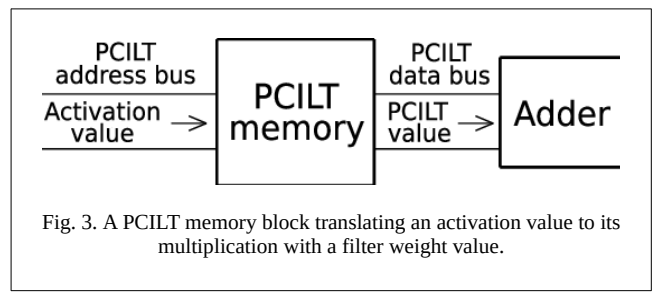

Fig. 3. A PCILT memory block translating an activation value to its multiplication with a filter weight value.

Thus, the convolution inference can be done by much smaller amount of circuitry than that in a modern CPU or GPU. Consequently, the on-chip area of an ASIC can house more such units than standard ALUs.

The inference speed bottleneck there will be the adder, having to process results from several PCILTs. Where the on-chip size is not critical, that might be sped up by having a tree of adders (Fig. 4).

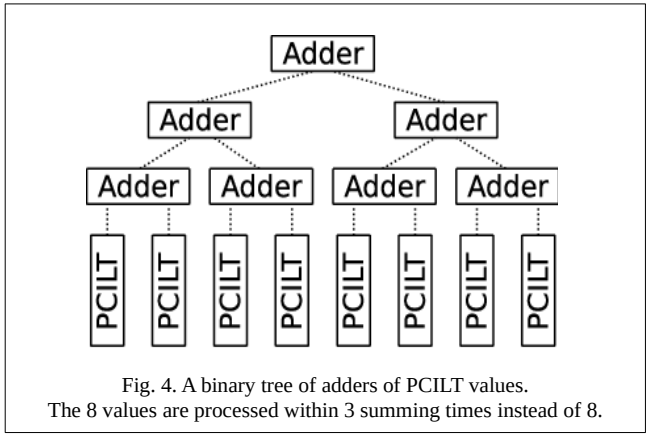

Fig. 4. A binary tree of adders of PCILT values. The 8 values are processed within 3 summing times instead of 8.

A CNN ASIC that uses the DM or separable convolutions algorithm does not need PCILT memory, but needs multiplication circuitry instead. Also, a multiplication by a memory value would be slower in it than an addressing with offset a value in fast (cached or static) memory. Thus, a DM-based ASIC is expected to be slower than a PCILT-based one.

CNN ASICs based on FFT or Toom-Cook algorithms can be theoretically faster than DM-using ones, but will need much more complex (and larger on-chip) circuitry, which will slow their execution and decrease the number of parallel units that can fit on a die. Preliminary estimates expect a PCILT implementation to be faster than them, too.

Creating the PCILTs adds an overhead to the CNN operation. However, it is done only once in the lifetime of a CNN, and will be negligible in most cases. For example, calculating the PCILTs for a 5x5 filter to process activations with 8-bit cardinality will require 6,400 multiplications. Processing with this filter 10,000 samples of size 1024x768 by DM will require 194,820,000,000 multiplications.

The main disadvantage of the PCILT algorithm is the amount of memory needed for the PCILTs. While investing in having more memory is often justified by the increased productivity over the system lifetime, the technical details can be problematic. In a modest-sized CNN – 5 convolutional layers, 50x80x120x200x350 neurons – using internally 8-bit activations and 5x5 filters with 8-bit values, PCILTs would need about 1.65 GB. If implemented as SRAM, fitting this amount of memory on a single die for maximal productivity will be costly, if possible at all at the technology level of 2021.

Lower activation cardinality can decrease significantly the PCILT memory. For example, 8-bit activations will need 256 values in a PCILT, while 4-bit activations will need only 16 values. With the latter, the example network mentioned above will need only about 100 MB for PCILTs.

Further on, the multiplication product of smaller-sized values can fit in less memory, decreasing the memory needed by a PCILT. The example network mentioned above thus can be shrunk to only about 75 MB for PCILTs. Several such networks, or a significantly larger one can fit on a single die, together with the other custom ASIC circuitry needed by the CNN work.

Also, a custom ASIC can have extractor PCILTs address bus with the activation bit width + PCILT number bit width, and data bus with the PCILT value bit width, thus potentially increasing the memory exchange speed and simplifying the processing circuitry. Such an ASIC likely will not be flexible enough for usage in research with diverse cases, but will be good for mass production applications.

PCILTs are calculated prior to the learning of the CNN, and do not change during the CNN lifetime. That allows storing them in ROM instead of RAM, on need.

In appropriate cases – eg. small PCILT tables / data cardinality, big data samples and hight speed priority – input weights can be calculated into the PCILT values, removing the need for input weights multiplication operation and for a multiplying ALU in the inference circuitry. (The multiplication operation will effectively be moved into the weight calculation and adjustment circuitry, where such an ALU will be present anyway.) This approach might increase the PCILT memory, and it might turn the weight adjustment into the speed bottleneck operation.

The PCILT values are an exact product of the convolutional function – there is no result precision loss. This makes the PCILT algorithm useful where both productivity and precision are important.

The algorithm works with both integer and FP weights of arbitrary size. Bigger sizes increase the speed advantage to the DM algorithm, but also need more PCILT memory. FP weights also increase the speed advantage over DM.

PCILTs allow productively utilizing inputs with different cardinalities – while calculating PCILT values, input data values cardinalities should be scaled to their lowest common denominator (LCD). If needed (eg. to save PCILT memory), even a max data value lower than the LCD can be used, at the cost of losing some precision from the inputs with the highest cardinality.

The PCILT algorithm is compatible with many other techniques for increasing performance – eg, with grouped convolutions. Obtaining results through PCILTS is usable well with some operations in separable convolutions. The algorithm extension *Using PCILTs as Weights* can also compensate for the parameter reduction in those.

PCILT values and even entire PCILTs can be optimized post-learning similarly to multiplication-using filter and input weights. Some quantization techniques might achieve better results with PCILTs than with most other CNNs data.

Post-learning, input weights can be incorporated into the PCILTs too, potentially at the cost of needing more memory.

Custom on-chip circuitry for this algorithm not only will be used much more sparsely than one for most other convolutional algorithms. It will also do less processing. So, its hardware implementation will need less cooling, or alternatively can be used more intensively.

## Extensions

This basic algorithm can be extended to improve some of its characteristics on certain systems and / or at certain tasks. Some extensions are described below.

### Pre-processing Activations Into PCILT Offsets

CNN layers receiving activations of small cardinality, running on systems with sufficient PCILT memory, can divide their filters into segments, each containing several weights. When processing a receptive field (RF), the activations matching a segment are combined into a single PCILT offset. The PCILTs in this case will contain not the convolutions of a single weight, but the sums of the convolutions of the weights in a segment (Fig. 5).

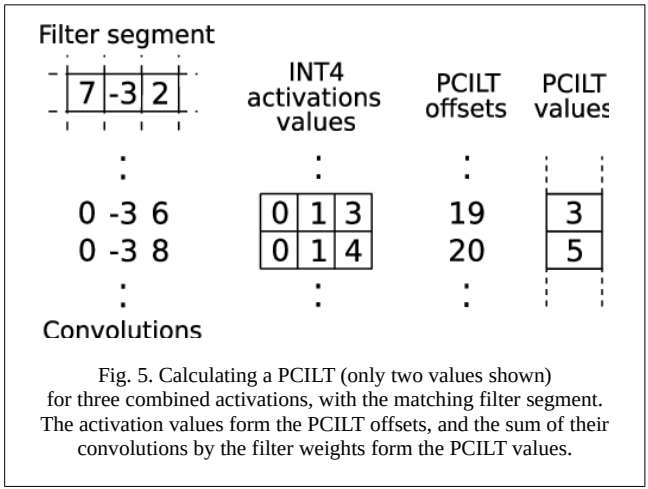

Fig. 5. Calculating a PCILT (only two values shown)
for three combined activations, with the matching filter segment.
The activation values form the PCILT offsets, and the sum of their
convolutions by the filter weights form the PCILT values.

In this way, the sum of the convolution results for an entire segment is retrieved as a single value (Fig. 6). This decreases both the count of PCILT memory accesses and the count of summed values, increasing the productivity. A test network with boolean activations, combining 8 activations into 8-bit offsets, achieved a speedup of 6.59x[73].

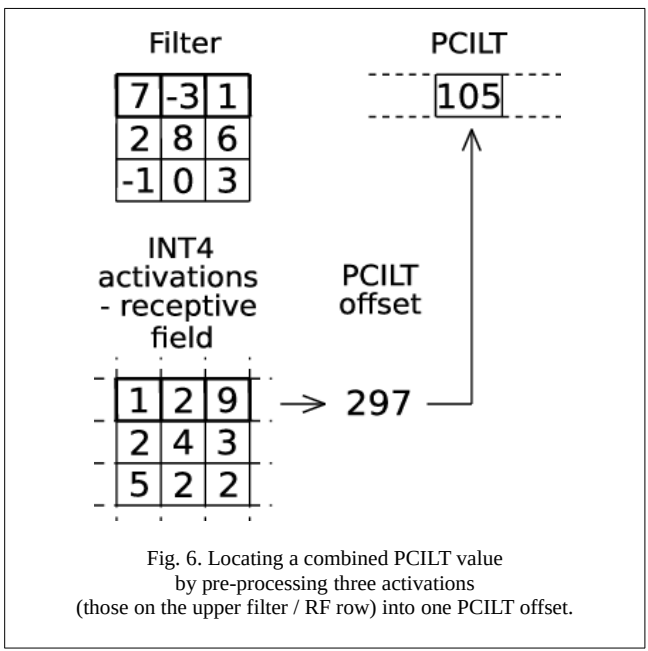

Fig. 6. Locating a combined PCILT value
by pre-processing three activations
(those on the upper filter / RF row) into one PCILT offset.

A custom ASIC can have activations data bus with the bit width of the activations combination / PCILT offset, thus increasing the speed and simplifying the data processing. An even wider data bus can extract several PCILT offsets at once, keeping them in a data register and shifting them one by one to the offset register, speeding additionally the activations memory accessing.

Activations can be pre-processed into PCILT offsets in more complex ways. In most cases this will require additional RAM for these offsets. However, it allows for combining non-adjacent activations, which can be used to various effects. For example, it allows for skipping some RF positions at all, thus eliminating non-important filter positions from being processed, and increasing the productivity. It also allows for including some RF positions in more than one PCILT – this can weigh them beyond the filter weights range, permitting the usage of a more limited range at the cost of a small delay (Fig. 7).

More complex combinations might increase the pre-processing overhead. Combining activations that are far apart in a cached memory (eg. ones in a big data sample, matching different filter rows) might increase the cache misses. Both these can decrease the productivity.

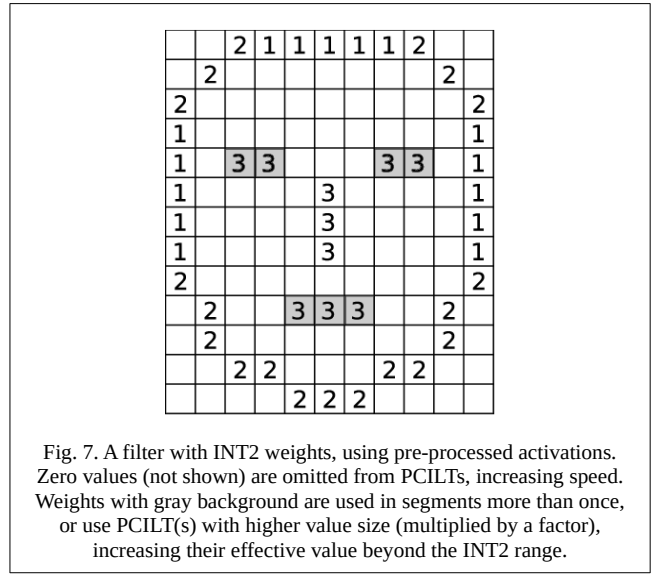

Fig. 7. A filter with INT2 weights, using pre-processed activations.
Zero values (not shown) are omitted from PCILTs, increasing speed.
Weights with gray background are used in segments more than once,
or use PCILT(s) with higher value size (multiplied by a factor),
increasing their effective value beyond the INT2 range.

The pre-processing can be done by separate circuitry, usually through fast operations (bit shifting and masking), pipelining the results to the convolutional circuitry. Thus, the overhead due to it can be minimal, esp. where the PCILT offsets are the same for the same inputs in different neurons, so calculated offsets can be reused.

In some cases, the PCILT offsets can be filter-specific. This might require separate combining circuitry and offsets memory for each filter.

There can also be hybrid approaches, where activations are combined in several different ways into data values, and some or all of the combination sets are used by several similar filters, possibly across different neurons. In some CNN architectures, PCILTs used in this process might be usable even to different layers.

A generalized version of this extension would see the activations as a bitstream that can be reprocessed into PCILT offsets in any needed way. For example, activations with bigger bit width can be split into smaller offsets, possibly also being combined in the process, thus decreasing the PCILT memory needed at the cost of some performance loss. This approach has a contiguous spectrum of trade-offs between memory economy and performance improvement, allowing for task/system-optimal balances between the two.

Different parts of activations can be reordered and merged into different offsets, locating their inference matches in different PCILTs, possibly calculated through different functions (see *Using Custom Convolutional Functions*). This can be used for eg. minimizing the number of unique PCILTs in CNNs with activations with many different bit sizes (see *Using Shared PCILTs*). Another usage might be achieving the effect of complex custom convolutional functions without sacrificing productivity (see also *Using Custom Convolutional Functions*).

### Using Custom Convolutional Functions

The classic convolutional operation is the multiplication of filter weights by activations. However, other functions can also be used, for example multiplying by logarithms or inverse logarithms of the filter weight and / or activation values. This can be used to re-scale and modify the range of the inferred values and their distribution through it, or to emulate the work of BNNs, etc.

One possible usage of this extension is using a range of integers to represent a bigger range of values with a non-uniform precision across it, eg. representing floating-point values with non-uniform distribution through integers with uniform distribution. This can increase the speed of processing in some cases.

Another usage might be the opposite – creating ranges of output values or internal activations values that reflect complex functions instead of simple uniform ranges. This can allow using internally small integer ranges (and thus applying the PCILT algorithm) where an output with a complex non-uniform range is needed.

Yet another usage is where a complex and / or slow convolutional function, or a set of different functions is preferable to the classic DM. Using PCILTs makes the extra computing power needed by this negligible in most cases. Their creation can be done by circuitry that is not a part of this ASIC, eg. by an external CPU or SOC, keeping the on-chip size of learning / working circuitry small.

Other usages might seek specific effects provided by the usage of custom convolutional functions, possibly aided by custom backpropagation weights adjustment. For example, this might be a fast way to achieve or approximate Bayesian convolution.

*Using Shared PCILTs*

PCILTs for the same convolutional algorithm base, eg. filter weight value(s), and activation cardinality are identical everywhere within a CNN. With PCILTs for different activations cardinalities, but with the same base, the one for the lower cardinality will match the beginning of the one for the higher cardinality. This allows to keep only one PCILT for given algorithm base value(s) and to replace the others with pointers to it.

This allows for limiting the amount of memory required for PCILTs. Often the weights in a filter might cover a big range of values (big overall cardinality), but might will use only a few of these (small actual cardinality), especially in some applications of extension *Pre-processing Activations Into PCILT Offsets*. The number of unique PCILTs in a CNN, or a part of it, is equal to the overall actual cardinality of its filter weights, multiplied by the number of the different activation cardinalities used there. A relatively liberal overall INT16 filter weights actual cardinality of 32, having 2 different activation cardinalities, INT10 and INT16, will require only about 25 MBs PCILT memory for an arbitrarily big CNN. If the filter values used are the same with the activations of different cardinalities, the PCILT memory can be decreased further to about 18 MB, as the PCILTs for the larger-cardinality activations will contain at their starts those for the smaller-cardinality activations.

Such an extension will require a more complex ASIC structure than the one shown in Fig. 3 – having a common PCILT memory and exchange bus, sharing of which may cause a processing delay. Another, smaller delay will come due to the usage of an additional PCILT indirection.

A case-specific balance between increasing the speed and decreasing the amount of memory used can be achieved by optimizing the bottleneck circuitry elements for parallel access and by increasing their repetitiveness.

A variant of this extension can use tables with indirection offsets to unique PCILT values instead of pointers to unique PCILTs. It can be useful where the repeatability among the entire PCILT tables is low, but that among the separate PCILT values is high. (This could be frequent where *Pre-processing Activations Into PCILT Offsets* is used, eg. to join into a PCILT offset several activations with very low cardinality. Together with filter weight values with relatively low actual cardinality, this can produce PCILTs that tend to contain the same values, but in different order.) It is feasible where the indirection offsets need substantially less memory than the PCILT values.

This variant might introduce some delay due to the additional indirection. However, a variant of it can provide for further memory reduction by grouping PCILT values with different useful precision in different shared PCILTs, each having values of different size. Where PCILT values precision allows for effective size equal to or smaller than that of the indirection tables values, they can be stored directly there, differing by a flag bit. This variant may be differently feasible on different platforms and systems.

In cases where the indirection offsets tables repeat often and the memory access speed is high, it might be justified to have two-level indirection: pointers to unique tables with indirection offsets to PCILTs with unique values.

This extension is mostly incompatible with *Using PCILTs as Weights*.

Some usages of *Pre-processing Activations Into PCILT Offsets*, specifically combining several activations into one offset, can increase the number of values in shared PCILTs $X^{(N-1)}$ times, where X is the actual filter weights cardinality, and N is the number of activations grouped into a PCILT offset. It might also increase the amount of memory a PCILT value needs.

*Using PCILTs as Weights*

This extension does not use input weights. Instead of them, during backpropagation it adjusts PCILT values, similarly to the CNNs that adjust filter weights instead of input weights. It brings a similarity to the BNNs which do not have segregation between pattern and input weights.

It eliminates the need to multiply the inference function dot result (IFDR) by an input weight, thus speeding up the inference, but potentially slowing the backpropagation.

The PCILT values can be product of any function, including ones that are not based on filter and input weights multiplication, or on a convolutional function at all. In an extreme case, they can even be generated randomly.

CNNs usually need more network parameters to achieve better precision. However, with almost all convolution algorithms that means also more computational load. This extension combines a big number of network parameters with the smaller computation load of the PCILTs, potentially solving this problem. The size of the PCILTs and their degree of sharing can be varied to achieve an optimal size of the network parameter space. The risk for needing more learning to achieve maximal precision, and the risk for slowing the backpropagation can be mitigated through appropriate weight adjustment algorithms. High repetitiveness of parameters can be regulated down by suitable network compression algorithms, post-learning or during learning.

There are four general ranges of adjusting PCILT values:

- changing all values in all PCILTs of a filter in the same way, effectively emulating the classic algorithm's multiplication of the IFDR by an input weight.
- changing all values in a PCILT in the same way, possibly accounting for the result of this specific PCILT. However, the PCILTs for different filter weights or weight groups can be changed in different ways. This is effectively equivalent to adjusting the filter weights in the classic DM algorithm.
- changing all same-offset values in all filter PCILTs in the same way, effectively equivalent to adjusting all filter weight values in the same way for the particular activation value that translates to this PCILT offset. However, for another activation value(s) the filter weight values might be adjusted in a different way. This is equivalent to effectively having different filter weights for different activations or activation combinations. This might be beneficial for some goals and detrimental for others.
- changing every value in a PCILT in a separate way, possibly accounting for the backpropagation result for the specific activation values translating to this PCILT value. This is equivalent to adjusting every filter weight specifically for every activation value. Potentially it allows for very flexible dependencies of the result by the incoming data, which can be beneficial or detrimental, depending on the goal. It might also need more training or goal-specific adjustment methods.

The more limited ranges are more selective, providing greater precision of affecting the results. They are also faster to execute. However, they also might need correspondingly more memory for preserving data of what activation values were convolved with what filter weights.

More selectivity can also bring abilities beyond these of a CNN with a single input weight per filter, eg. creating during the learning phase complex non-linear results for every combination of filter weight value and activation value. These are achieved by a relatively simple on-chip circuitry, giving speed, on-chip size and energy consumption advantages. However, more selectivity can also be detrimental in some cases. For example, it might slow down the learning in the CNN region of its application, and to slow down, skew or even disrupt the learning in the layers above it.

Custom ranges and / or non-linear adjustments within a range can be used too, seeking optimal trade-offs between results precision, processing speed, memory usage and energy consumption, and / or goal specificity.

If the extension *Pre-processing Values Into PCILT Offsets* is used too, there can be (except in the full range) multiple sub-variants for determining a PCILT changes base and a range of PCILTs affected by this change.

When used with *Pre-processing Activations Into PCILT Offsets*, this extension allows for reprocessing together activations from different inputs, creating neuron-wide filters. (Normally this would be impossible due to the need to weight the IFDR by input-specific weights.) These can even be shared by different neurons, turning effectively into layer-wide filters. This can increase the productivity by decreasing calculations repeatability. It can also provide some benefits while designing CNNs and their layers, for example introducing desired dependencies between different filters in a layer, or even different neurons in a filter, that can perform very fast.

At the end of the training, it might be possible to analyze the final PCILT values and to build back from them weight-adjusted input filters. These can be used in testing / work conditions with the classic DM algorithm, avoiding the need to multiply the IFDR by an input weight. The breakdown of the PCILT values by filter weight value and activation value might provide for better results at some CNN optimization methods. The other approach is also possible.

One of the applications for this extension might be increasing the speed in cases where the PCILT values in the zone of application are substantially fewer than the values inferred there, eg. with small PCILTs and big data samples.

Another application might be the utilizing of the extra abilities this extension provides – higher result selectivity, neuron-wide and layer-wide filters, etc.

Another application might be in systems with extremely tight memory consumption requirements, in work cases where this approach allows to omit energy-intensive circuitry, eg. some of the multiplying ALUs.

Yet another application can be in already learned systems – they do not change their input weights, thus not having this backpropagation overhead. These might achieve very high processing speed, esp. in highly specific custom ASICs where the PCILT memory is tightly integrated with the processing circuitry.

A disadvantage of this extension is using much memory, both for PCILTs and for storing temporary data during learning.

This extension is mostly incompatible with *Using Shared PCILTs*. Depending of the backpropagation implementation, using slow custom convolutional functions might decrease the productivity.